# Network Traffic Classification Using Machine Learning, Transformer, and Large Language Models

Ahmad Antari
*Security Department, Blue-Team*
*JAWWAL*
Nablus, Palestine
a.antari2@student.aaup.edu

Yazan Abo-Aisheh
*Technology Systems Department*
*Palestine Monetary Authority*
Ramallah, Palestine
y.abuaisheh@student.aaup.edu

Jehad Shamasneh
*Department of Natural, Engineering and Technology Sciences, Faculty of Graduate Studies*
*Arab American University*
Ramallah, Palestine
j.shamasneh@student.aaup.edu

Huthaifa I. Ashqar
*Civil Engineering Department*
*Arab American University*
Jenin, Palestine
huthaifa.ashqar@aaup.edu

*Abstract*—This study uses various models to address network traffic classification, categorizing traffic into web, browsing, IPSec, backup, and email. We collected a comprehensive dataset from Arbor Edge Defender (AED) devices, comprising of 30,959 observations and 19 features. Multiple models were evaluated, including Naive Bayes, Decision Tree, Random Forest, Gradient Boosting, XGBoost, Deep Neural Networks (DNN), Transformer, and two Large Language Models (LLMs) including GPT-4o and Gemini with zero- and few-shot learning. Transformer and XGBoost showed the best performance, achieving the highest accuracy of 98.95 and 97.56%, respectively. GPT-4o and Gemini showed promising results with few-shot learning, improving accuracy significantly from initial zero-shot performance. While Gemini Few-Shot and GPT-4o Few-Shot performed well in categories like Web and Email, misclassifications occurred in more complex categories like IPSec and Backup. The study highlights the importance of model selection, fine-tuning, and the balance between training data size and model complexity for achieving reliable classification results.

I. INTRODUCTION

The expansion of internet usage and the complexity of online applications have made the classification of internet traffic an essential aspect of network management. It is important also for many applications, covering the range from Quality-of-Service (QoS) provisioning to network security to resource management. Port-based and payload inspection techniques and other traditional traffic classification methods has been proved to be very limited due to port obfuscation, encryption and the dynamic behavior of modern internet applications.

Internet traffic classification has highly benefited from machine learning (ML) techniques demonstrated by their ability to discover patterns and classify with high precision. Unlike conventional approaches that rely on predefined rules and handcrafted features, ML methods can learn about new types of traffic as well as changing network conditions [1]. Also, Deep Learning (DL) techniques have been successful in dealing with intricate and multiplex modern network traffic by automatically selecting the appropriate attributes during training [1].

This paper explores various approaches including ML, DL, transformers and Large Language Models (LLMs) to internet traffic classification. We present a comprehensive review of existing methodologies and introduce a new framework that integrates feature extraction and classification into a single system. Additionally, we address the challenges posed by encrypted traffic and discuss the benefits of using different approaches to overcome these obstacles.

### A. Problem Statement and Objectives

It is quite challenging for traditional traffic classification models to handle the rising internet usage because of the increased use of encrypted communication and the increasing number of connected devices to the web. With port obfuscation being widely adopted, encryption, and dynamic modern-day web applications, port-based techniques and payload inspection are no longer useful. They are also unsuitable for efficient network management, QoS provisioning, accurate billing, and robust intrusion detection systems. Therefore, there is a huge demand for advanced flexible, and accurate traffic classification approaches to overcome these limitations and ensure the effective operation and security of contemporary networks. The aim of this study is to 1) Critically assess different machine learning models used in network traffic classification, 2) Contrast cutting-edge techniques like transformers and LLMs in network traffic classification, 3) Create a resilient framework that combines feature extraction with classification to improve the accuracy and efficiency of traffic classification in various network environments, and 4) Collect a new dataset from AED that includes web, browsing, IPSec, backup, and email.

II. LITERATURE REVIEW

Network traffic classification involves categorizing network traffic into different classes to enhance various network management and security applications such as QoS provisioning, billing, and intrusion detection [2], [3], [4]. Traditional methods such as port-based and payload inspection have limitations due to the increasing use of encrypted traffic and dynamic port assignments. Machine learning and deep learning methods have emerged as effective solutions for traffic classification, offering higher accuracy and adaptability to evolving traffic patterns [5], [6].

The data sources for network traffic classification vary widely across different studies. Lotfollahi used encrypted traffic datasets to evaluate their deep learning approach [1], while Sarhan focused on NetFlow datasets derived from four benchmark NIDS datasets: UNSW-NB15, BoT-IoT, ToN-IoT, and CSE-CIC-IDS2018 [5]. Shafiq utilized various smart city traffic datasets, emphasizing the importance of diverse and representative data [7]. Menezes and Mello employed flow feature-based datasets collected using tools like Wireshark and Argus [8]. Churcher analyzed IoT network traffic datasets to classify attacks [9]. Pacheco reviewed multiple datasets used in deploying ML solutions for traffic classification, highlighting the need for standardization [10]. Nguyen and Armitage used datasets such as the ISCX VPN-nonVPN and Moore datasets [11]. Rezaei and Liu provided an overview of deep learning applications in traffic classification using encrypted traffic datasets [12]. Wang converted existing datasets into NetFlow format for their study [13].

The results of network traffic classification studies also vary depending on the methodologies and datasets used. Nguyen and Armitage found that SVM and decision trees provided the highest accuracy in their comparative study [11]. Huang achieved significant improvements in classification accuracy using statistical feature-based methods [14]. Yuan achieved 94.2% accuracy using SVM for internet traffic classification [15]. Wang demonstrated high accuracy using a token-based approach [16]. Singh reported high accuracy in near real-time classification using ML techniques [17]. Menezes achieved approximately 90% accuracy using flow feature-based methods with KNN [8]. Churcher reported 92.8% accuracy in IoT attack classification using ML algorithms [9]. Kumar achieved 92.8% accuracy in IoT traffic classification [18].

The advancements in machine learning and deep learning methodologies have significantly improved the accuracy and efficiency of network traffic classification. Traditional methods like port-based and payload-based techniques have been largely superseded by statistical feature-based and deep learning methods due to their ability to handle encrypted traffic and dynamic port usage. The development of comprehensive datasets and the use of sophisticated models like CNNs and RNNs have demonstrated the potential for highly accurate traffic classification in various network environments. Future research should focus on refining these methods and exploring new techniques to further enhance the robustness and adaptability of network traffic classification systems. Additionally, efforts should be made to standardize datasets and features used for training and evaluating advanced classification models including LLMs [19], [20].

Our primary contribution lies in creating a new dataset and the extensive comparison of various ML models, and advanced techniques like transformers and LLMs for network traffic classification. We evaluated multiple algorithms including Naive Bayes, Decision Tree, Random Forest, Gradient Boosting, XGBoost, and Deep Neural Networks (DNN), each tuned for optimal performance. Additionally, we explored the efficacy of transformer models and LLMs such as Gemini and GPT-4o in enhancing classification accuracy. In addition to highlighting the advantages and disadvantages of each technique, this thorough comparison offers insightful information on how these models might be used in actual network traffic classification situations.

III. METHODOLOGY

The diagram below shows the methodology that will be followed in seeking a solution for classifying network traffic.

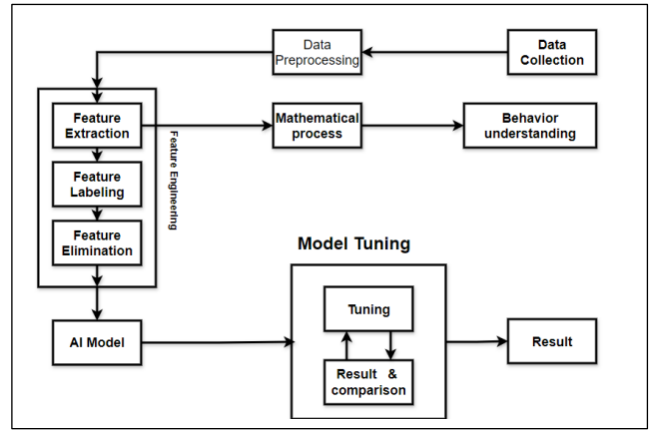

Fig. 1. Proposed methodology flowchart.

A. Dataset Description

The dataset was collected from AED (Arbor Edge Defender) devices as pcap files each one containing 5,000 packets, after that by applying Python script to each file that converts the pcap file into csv one, and the packets to flows, then we combined them to have one dataset that consists of 30,959 rows and 19 columns with five classes, the features are shown in TABLE I. The classes were chosen regarding the organization's top services and here we should mention the difference between the web and browsing, the web is related to the traffic from the servers which correlated to the online services and web pages, number of rows (flows) per class, and the ratios are shown in TABLE II.

TABLE I. FEATURES DESCRIPTION.

| Feature | Brief Description |
|---|---|
| flow_id | Identifier for the flow |
| flow_ip_src | The source IP address of the flow |
| flow_ip_dst | The destination IP address of the flow |
| flow_srcport | Source port of the flow |
| flow_dstport | The destination port of the flow |
| flow_proto | Protocol used in the flow (e.g., TCP, UDP) |
| num_packets | Number of packets in the flow |
| total_length | The total length of the packets in the flow |
| avg_packet_size | The average size of packets in the flow |
| min_time | Minimum timestamp of the flow |

| max_time | Maximum timestamp of the flow |
|---|---|
| tcp_window_size_avg | Average TCP window size in the flow |
| total_payload | Total payload size of the flow |
| forward_packets | Number of packets forwarded in the flow |
| receiving_packets | Number of packets received in the flow |
| fragments | Number of fragmented packets in the flow |
| flow_duration | Duration of the flow |
| Target | Target class or label for classification |
| Target as numeric | Numeric representation of the target class |

TABLE II. MODEL CLASSES.

| Class # | Class | Count | Percentage |
|---|---|---|---|
| 0 | Backup | 6,444 | 20.82% |
| 1 | IPSec | 6,349 | 20.51% |
| 2 | Browsing | 6,135 | 19.82% |
| 3 | Web | 6,016 | 19.43% |
| 4 | Email | 6,015 | 19.43% |

### B. Feature Engineering

Random Forest Machine learning model was used to draw the feature importance plot as shown in the Fig. 2 and Fig. 3. Based on this, we deleted the "min_time" and "max_time" features to avoid model biasing taking into consideration that these features are represented in another variable called "Duration". Finally, the 'fragments' feature was dropped from the data set because it has one value, 'forward_packets', 'receiving_packets' represented in another way as 'num_packets', 'flow_proto' features failed in the p-value test. so, they were eliminated, which is shown in Fig. 3.

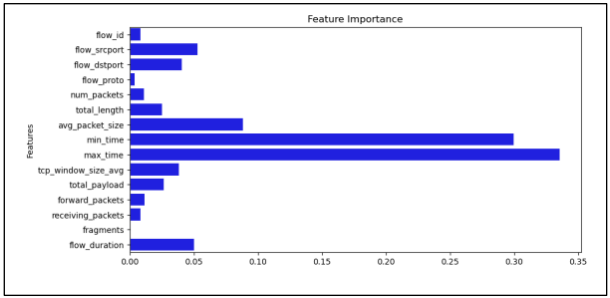

Fig. 2. Feature importance using RF with all features.

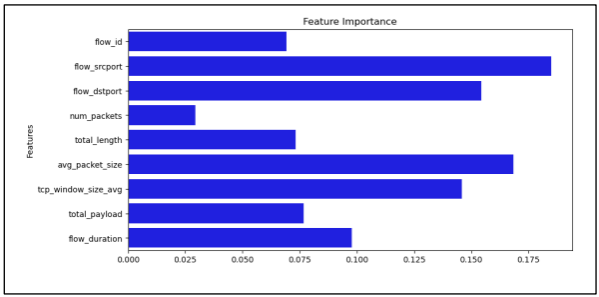

Fig. 3. Feature importance using RF Model after the Elimination Process.

Fig. 4 shows the correlation between the features, noted that there was a correlation between the features 'total_payload' and 'total_length' after checking the dataset, we found that 96% of the values are equal, and by testing which one has less effect on the accuracy we dropped the feature 'total_length'.

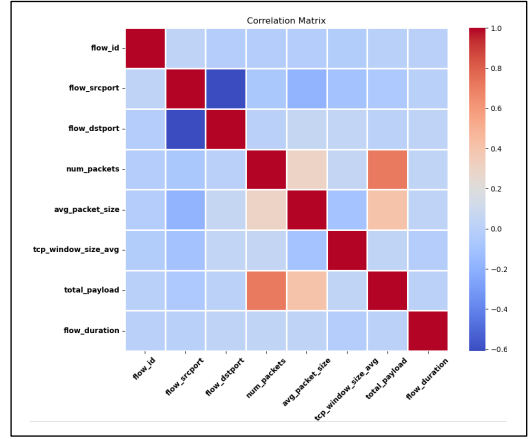

Fig. 4. Correlation Matrix.

### C. Proposed Classification Models

Different Machine learning models were used in this study including Naive Bayes, Multinomial Naive Bayes, Random Forest, Decision tree, Gradient Boosting, XGBoost, and Deep Neural network. the results from these models went through a tuning process using five cross-validations and a grid search for hyperparameter tuning. Finally, the results from all models were compared with the same conditions and the best model was selected. We employed AutoGluon Tabular Predictor with a Transformer-based architecture as Auto ML. The model was trained on the provided dataset after completing the necessary preprocessing steps, such as min-max scaling for numerical features and label encoding for categorical features. The hyperparameter search strategy was selected as 'best_quality' to prioritize accuracy over training time. On the Other hand, we used LLMs as zero-shot and few-shot for GPT-4o and Gemini.

## IV. RESULTS AND DISCUSSION

TABLE III shows the overall results for the used models. All models were tested using the train-test ratio of 80-20 as it was the best ratio for the performance. TABLE III provides a comparison of different models across accuracy, accuracy after tuning, and F1-score. It highlights the impact of tuning on model performance and the relative strengths of traditional, ensemble, and advanced deep learning methods. Traditional models like Naive Bayes and Multinomial Naive Bayes exhibit modest performance. Naive Bayes achieves an accuracy of 34.51%, improving slightly to 35.76% after tuning, with an F1-score of 33.58%. Similarly, Multinomial Naive Bayes starts at 36.06% accuracy and moves to 36.13%, with an F1-score of 35.82%, indicating only marginal benefits from tuning.

Tree-based models perform significantly better. Decision Trees start with an impressive accuracy of 95.46%, but tuning causes a slight dip to 95.00%, and they maintain an F1-score of 92.56%. Random Forest achieves 96.71% accuracy both before and after tuning, with an F1-score of 97.34%, showcasing its robustness. Gradient Boosting and XGBoost stand out with their high performance. Gradient Boosting improves from 91.93% to 96.01% accuracy after tuning, with an F1-score of 95.30%. XGBoost, however, leads among ensemble methods, improving from 97.20% to 97.56% accuracy and achieving an F1-score of 98.39%.

Deep learning models show notable trends. Deep Neural Networks exhibit a dramatic leap from 46.44% accuracy to 86.04% after tuning, with an F1-score of 85.44%, highlighting the transformative impact of tuning. Transformers outperform all other models, with accuracy improving from 97.70% to 98.95%, and an F1-score of 98.82%, setting the benchmark for performance. However, large language models like GPT-4 and Gemini demonstrate interesting patterns in zero-shot and few-shot settings. GPT-4 achieves 31.40% accuracy in zero-shot but improves significantly to 61.40% in few-shot, with corresponding F1-scores of 20.27% and 60.56%. Similarly, Gemini achieves 41.14% accuracy in zero-shot and 69.80% in few-shot, with F1-scores of 40.06% and 68.91%, illustrating the effectiveness of few-shot learning.

Results show that advanced models like Transformers and XGBoost dominate in both accuracy and F1-scores. Tuning significantly benefits certain models, especially deep learning-based approaches, while few-shot learning proves essential for extracting higher performance from large language models.

TABLE III. RESULTS OF USED MODELS.

| ML Model | Accuracy | Accuracy After Tuning | F1-Score |
|---|---|---|---|
| Naive Bayes | 34.51 | 35.76 | 33.58 |
| Multinomial Naive Bayes | 36.06 | 36.13 | 35.82 |
| Decision Tree (DT) | 95.46 | 95.00 | 92.56 |
| Random Forest | 96.71 | 96.71 | 97.34 |
| Gradient Boosting | 91.93 | 96.01 | 95.30 |
| XGboost | 97.20 | 97.56 | 98.39 |
| Deep Neural Network | 46.44 | 86.04 | 85.44 |
| Transformer | 97.70 | 98.95 | 98.82 |
| GPT-4o Zero-Shot | 31.40 | - | 20.27 |
| GPT-4o Few-Shot | 61.40 | - | 60.56 |
| Gemini Zero-Shot | 41.14 | - | 40.06 |
| Gemini Few-Shot | 69.80 | - | 68.91 |

### A. LLMs Confusion Matrices

This section presents the performance of different models (Gemini Zero-Shot, GPT-4o Zero-Shot, Gemini Few-Shot, and GPT-4o Few-Shot) across five categories: Backup, Browsing, Email, IPSec, and Web. These matrices highlight the models' strengths and weaknesses in correctly classifying instances and offer insights into misclassifications, allowing for a better understanding of each model's effectiveness in various tasks.

The Gemini Zero-Shot confusion matrix in TABLE IV shows mixed performance across categories. It performs well with Browsing and Email, correctly predicting 37 out of 70 and 54 out of 147 instances, respectively. However, the model struggles with Backup (19 out of 49 correctly predicted) and Web (7 out of 55), where significant misclassifications occur, especially with Web being confused with Email (44 instances). The GPT-4o Zero-Shot confusion matrix in TABLE V shows high misclassification, especially with Web, where 91 out of 280 instances are correctly predicted, but 60 instances are misclassified as Browsing. Backup and Email also face significant confusion, with 66 and 79 instances misclassified, respectively. However, Web is correctly predicted in 91 instances, showing the best performance.

TABLE IV. GEMINI ZERO-SHOT CONFUSION MATRIX.

| | | Predicted | | | | |
|---|---|---|---|---|---|---|
| | | Backup | Browsing | Email | IPSec | Web |
| True Label | Backup | 19 | 12 | 36 | 0 | 3 |
| | Browsing | 1 | 37 | 13 | 0 | 19 |
| | Email | 3 | 5 | 54 | 0 | 8 |
| | IPSec | 19 | 6 | 0 | 27 | 18 |
| | Web | 7 | 10 | 44 | 2 | 7 |
| | Total | 49 | 70 | 147 | 29 | 55 |

TABLE V. GPT-4O ZERO-SHOT CONFUSION MATRIX.

| | | Predicted | | | | |
|---|---|---|---|---|---|---|
| | | Backup | Browsing | Email | IPSec | Web |
| True Label | Backup | 66 | 16 | 0 | 0 | 18 |
| | Browsing | 40 | 0 | 0 | 0 | 60 |
| | Email | 1 | 20 | 0 | 0 | 79 |
| | IPSec | 44 | 24 | 0 | 0 | 32 |
| | Web | 1 | 7 | 1 | 0 | 91 |
| | Total | 152 | 67 | 1 | 0 | 280 |

The Gemini Few-Shot confusion matrix in TABLE VI shows good performance, especially for Web, where 82 out of 112 instances are correctly predicted. Browsing also performs well, with 73 out of 103 instances correctly classified. However, misclassification occurs in Backup (67 correct out of 107) and Email (66 correct out of 85), where some instances are misclassified as other categories. The GPT-4o Few-Shot confusion matrix in TABLE VII shows strong performance for Email, with 100 out of 203 instances correctly predicted. Browsing and Backup also show good results, with 69 correct predictions for each. However, there is significant misclassification in IPSec, with 43 instances misclassified as Browsing, and Web, where 69 instances are misclassified as Email.

TABLE VI. GEMINI FEW-SHOT CONFUSION MATRIX.

| | | Predictive | | | | |
|---|---|---|---|---|---|---|
| | | Backup | Browsing | Email | IPSec | Web |
| True Label | Backup | 67 | 10 | 3 | 13 | 7 |
| | Browsing | 14 | 73 | 4 | 9 | 0 |
| | Email | 8 | 14 | 66 | 1 | 11 |
| | IPSec | 17 | 1 | 9 | 61 | 12 |
| | Web | 1 | 5 | 3 | 9 | 82 |
| | Total | 107 | 103 | 85 | 93 | 112 |

TABLE VII. GPT-4O FEW-SHOT CONFUSION MATRIX.

| | | Predictive | | | | |
|---|---|---|---|---|---|---|
| | | Backup | Browsing | Email | IPSec | Web |

| True Label | | | | | | |
|---|---|---|---|---|---|---|
| | Backup | 69 | 1 | 13 | 17 | |
| | Browsing | 7 | 69 | 19 | 0 | 5 |
| | Email | 0 | 0 | 100 | 0 | 0 |
| | IPSec | 0 | 43 | 2 | 43 | 12 |
| | Web | 0 | 5 | 69 | 0 | 26 |
| | Total | 76 | 118 | 203 | 60 | 43 |

The results from the confusion matrices highlight varying levels of performance across the different LLM models. Gemini Few-Shot and GPT-4o Few-Shot show strong classification in certain categories, particularly Web and Email, but also face misclassifications, especially in more complex categories like IPSec and Backup. The zero-shot models, Gemini and GPT-4o, exhibit higher misclassification rates, particularly in specific categories like Web and Browsing. These findings underscore the models' ability to handle simple tasks with fewer examples, while also emphasizing the need for improvement in handling more intricate class distinctions. The results demonstrate that while advanced models like Gemini and GPT-4o offer promising performance, fine-tuning and further training are essential for optimizing accuracy across diverse categories.

## V. CONCLUSION

In this study, we addressed the problem of network traffic classification by categorizing traffic into web, browsing, IPSec, backup, and email using various models. Our dataset, collected from Arbor Edge Defender (AED) devices, enabled a thorough analysis and evaluation of multiple algorithms, including Naive Bayes, Decision Tree, Random Forest, Gradient Boosting, XGBoost, and Deep Neural Networks (DNN), Transformer, and LLMs. Among the evaluated models, Transformer and XGBoost showed the best performance, achieving the highest accuracy of 98.95 and 97.56%, respectively. Their ability to handle different feature scales and complex relationships was key to these results. Additionally, LLMs like Gemini and GPT-4o demonstrated potential, especially with few-shot learning, though they required more fine-tuning and computational resources.

Results show that while advanced models like Gemini Few-Shot and GPT-4o Few-Shot exhibit strong performance in certain categories, such as Web and Email, they also encounter significant misclassifications, especially in more complex categories like IPSec and Backup. The zero-shot models demonstrate weaker overall performance, with particularly high misclassification rates in categories like Web and Browsing. Challenges remain in achieving optimal performance across all categories, indicating the need for further tuning and training.

REFERENCES

[1] M. Lotfollahi, M. Jafari Siavoshani, R. Shirali Hossein Zade, and M. Saberian, "Deep packet: a novel approach for encrypted traffic classification using deep learning," *Soft comput*, vol. 24, no. 3, pp. 1999–2012, Feb. 2020, doi: 10.1007/s00500-019-04030-2.
[2] H. Abu-Helo and H. I. Ashqar, "Early Ransomware Detection System Based on Network Behavior," 2024. doi: 10.1007/978-3-031-57931-8_43.
[3] T. T. T. Nguyen and G. Armitage, "A survey of techniques for internet traffic classification using machine learning," *IEEE communications surveys & tutorials*, vol. 10, no. 4, pp. 56–76, 2008.
[4] A. Weshahi, F. Dwaik, M. Khouli, H. I. Ashqar, A. M. J. Shatnawi, and M. Elkhodr, "IoT-Enhanced Malicious URL Detection Using Machine Learning," 2024. doi: 10.1007/978-3-031-57931-8_45.
[5] M. Sarhan, S. Layeghy, N. Moustafa, and M. Portmann, "NetFlow Datasets for Machine Learning-based Network Intrusion Detection Systems," Nov. 2020, doi: 10.1007/978-3-030-72802-1_9.
[6] Y. M. Tashtoush et al., "Agile Approaches for Cybersecurity Systems, IoT and Intelligent Transportation," *IEEE Access*, vol. 10, pp. 1360–1375, 2021.
[7] M. Shafiq, Z. Tian, A. K. Bashir, A. Jolfaei, and X. Yu, "Data mining and machine learning methods for sustainable smart cities traffic classification: A survey," *Sustain Cities Soc*, vol. 60, p. 102177, Sep. 2020, doi: 10.1016/j.scs.2020.102177.
[8] N. A. T. de Menezes and F. L. de Mello, "Flow Feature-Based Network Traffic Classification Using Machine Learning," *Journal of Information Security and Cryptography (Enigma)*, vol. 8, no. 1, pp. 12–16, Dec. 2021, doi: 10.17648/jisc.v8i1.79.
[9] A. Churcher et al., "An Experimental Analysis of Attack Classification Using Machine Learning in IoT Networks," *Sensors*, vol. 21, no. 2, p. 446, Jan. 2021, doi: 10.3390/s21020446.
[10] F. Pacheco, E. Exposito, M. Gineste, C. Baudoin, and J. Aguilar, "Towards the Deployment of Machine Learning Solutions in Network Traffic Classification: A Systematic Survey," *IEEE Communications Surveys & Tutorials*, vol. 21, no. 2, pp. 1988–2014, Oct. 2019, doi: 10.1109/COMST.2018.2883147.
[11] T. T. T. Nguyen and G. Armitage, "A survey of techniques for internet traffic classification using machine learning," *IEEE Communications Surveys & Tutorials*, vol. 10, no. 4, pp. 56–76, Dec. 2008, doi: 10.1109/SURV.2008.080406.
[12] S. Rezaei and X. Liu, "Deep Learning for Encrypted Traffic Classification: An Overview," *IEEE Communications Magazine*, vol. 57, no. 5, pp. 76–81, May 2019, doi: 10.1109/MCOM.2019.1800819.
[13] Y. Wang, Y. Xiang, J. Zhang, W. Zhou, G. Wei, and L. T. Yang, "Internet Traffic Classification Using Constrained Clustering," *IEEE Transactions on Parallel and Distributed Systems*, vol. 25, no. 11, pp. 2932–2943, Nov. 2014, doi: 10.1109/TPDS.2013.307.
[14] Shijun Huang, Kai Chen, Chao Liu, A. Liang, and Haibing Guan, "A statistical-feature-based approach to internet traffic classification using Machine Learning," in *2009 International Conference on Ultra Modern Telecommunications & Workshops*, IEEE, Oct. 2009, pp. 1–6. doi: 10.1109/ICUMT.2009.5345539.
[15] R. Yuan, Z. Li, X. Guan, and L. Xu, "An SVM-based machine learning method for accurate internet traffic classification," *Information Systems Frontiers*, vol. 12, no. 2, pp. 149–156, Apr. 2010, doi: 10.1007/s10796-008-9131-2.
[16] T. Bakhshi and B. Ghita, "On Internet Traffic Classification: A Two-Phased Machine Learning Approach," *Journal of Computer Networks and Communications*, vol. 2016, pp. 1–21, 2016, doi: 10.1155/2016/2048302.
[17] K. Singh, S. Agrawal, and B. S. Sohi, "A Near Real-time IP Traffic Classification Using Machine Learning," *International Journal of Intelligent Systems and Applications*, vol. 5, no. 3, pp. 83–93, Feb. 2013, doi: 10.5815/ijisa.2013.03.09.
[18] R. Kumar, M. Swarnkar, G. Singal, and N. Kumar, "IoT Network Traffic Classification Using Machine Learning Algorithms: An Experimental Analysis," *IEEE Internet Things J*, vol. 9, no. 2, pp. 989–1008, Jan. 2022, doi: 10.1109/JIOT.2021.3121517.
[19] A. Radwan, M. Amarneh, H. Alawneh, H. I. Ashqar, A. AlSobeh, and A. A. A. R. Magableh, "Predictive Analytics in Mental Health Leveraging LLM Embeddings and Machine Learning Models for Social Media Analysis," *International Journal of Web Services Research (IJWSR)*, vol. 21, no. 1, pp. 1–22, 2024.
[20] H. I. Ashqar, A. Jaber, T. I. Alhadidi, and M. Elhenawy, "Advancing Object Detection in Transportation with Multimodal Large Language Models (MLLMs): A Comprehensive Review and Empirical Testing," *arXiv preprint arXiv:2409.18286*, 2024.